\documentclass[letterpaper, 10 pt, conference]{ieeeconf}  % Comment this line out if you need a4paper

\IEEEoverridecommandlockouts                              % This command is only needed if 
\usepackage{kotex} 
\usepackage{bm}
\usepackage{amsmath}
\usepackage{amssymb}
\usepackage{algorithm}
\usepackage{algpseudocode}
\usepackage{graphicx}
\usepackage{url}
\usepackage{tabularx}

\title{\LARGE \bf
Scale--Invariant Manipulability Shape Tracking Across Heterogeneous Manipulators
}

\author{
Geunwoo Kwon$^{1}$,
Dong-gyu Lee$^{1}$,
Kai Li$^{1}$,
Soonwoong Hwang$^{2}$,
and Wansoo Kim$^{2,\dagger}$
\thanks{$^\dagger$ Corresponding author.}
\thanks{$^{1}$G. Kwon, D. Lee and K. Li are with Hanyang University,
Seoul, Republic of Korea
(e-mail: {\tt\small gunwoo00@hanyang.ac.kr}).}
\thanks{$^{2}$S. Hwang and W. Kim are with Hanyang University ERICA,
Ansan, Republic of Korea
(e-mail: {\tt\small wansookim@hanyang.ac.kr}).}
}

\begin{document}

\maketitle
\thispagestyle{empty}
\pagestyle{empty}

%%%%%%%%%%%%%%%%%%%%%%%%%%%%%%%%%%%%%%%%%%%%%%%%%%%%%%%%%%%%%%%%%%%%%%%%%%%%%%%%
\begin{abstract}
When transferring manipulability across systems with different sizes and kinematic structures, matching absolute ellipsoid scale may be unnecessary when the goal is to reproduce orientation and semi-axis length ratios.
Full-matrix tracking, however, penalizes both shape and absolute-scale differences, even when only shape matching is required.
We therefore propose a scale-invariant manipulability
shape-tracking method that treats matrices differing only
by a positive scalar factor as equivalent and uses their
unit-determinant representatives.
We derive the differential of the unit-determinant shape representative and an orthonormal coordinate representation of the tangent tracking residual under the affine-invariant Riemannian metric (AIRM).
The resulting scale-invariant objective is integrated with position and end-effector direction tasks in a constrained joint-velocity quadratic program.
Simulations with four heterogeneous robots evaluate robot-to-robot and human-to-robot transfer.
On three followers, the proposed method achieves endpoint shape distances of $9.30\times10^{-5}$ without scale tuning.
With robot-specific target scales tuned during motion, the Full method retains endpoint axis-ratio errors of $0.19$--$0.31$ on KR500 and UR20.
For human reaching with concurrent tasks, the proposed method yields dual force shapes elongated along $X$ like the human target on all four robots, with endpoint position errors of $2.4$--$5.6\%$ of reference arm length versus up to $75\%$ for the Full method tracking the original human ellipsoid.
\end{abstract}

%%%%%%%%%%%%%%%%%%%%%%%%%%%%%%%%%%%%%%%%%%%%%%%%%%%%%%%%%%%%%%%%%%%%%%%%%%%%%%%%

\section{Introduction}

Selecting robot postures for manipulation requires considering
not only end-effector (EE) trajectories but also motion and force
transmission capabilities in task-relevant directions
\cite{chiu1988task,ajoudani2017choosing}.
Under a given kinematic model and prescribed Euclidean
joint-velocity or joint-torque norm constraints, a manipulability
ellipsoid (ME) represents these capabilities through its
principal directions and semi-axis lengths
\cite{yoshikawa1985manipulability}.
Semi-axis lengths describe achievable magnitudes along the
principal directions, and their ratios capture the relative
directional structure.
In their analysis of human screwing movements, Jaquier et al.
reported phase-dependent changes in velocity ME geometry
that were not fully captured by the determinant and condition
number \cite{jaquier2020analysis}.
These observations motivate using ME orientation and semi-axis
length ratios to describe and transfer posture-dependent
directional capabilities.

Previous studies have specified desired manipulability
geometry from task requirements
\cite{jaquier2018geometry,yan2021decentralized,choi2023simple}
or learned manipulability profiles from demonstrations for
reproduction on another robot
\cite{rozo2017learning,jaquier2021geometry}.
Learned profiles have also been adapted to new intermediate
manipulability targets \cite{abu2021probabilistic}
and reproduced using geometry-aware dynamic movement
primitives \cite{abu2024unified}.
More recently, ManiDP incorporated manipulability-based
guidance into a diffusion policy trained on teleoperated
demonstrations for bimanual manipulation
\cite{li2025manidp}.
For tracking a prescribed ME profile, Jaquier et al.
treated manipulability matrices as points on the manifold
of symmetric positive-definite (SPD) matrices and developed
geometry-aware tracking based on the Riemannian logarithmic
error between the current and reference MEs
\cite{jaquier2021geometry}.
Because the reference specifies the complete SPD matrix,
this full-matrix tracking includes differences in orientation,
semi-axis length ratios, and absolute scale in the tracking
error.

\begin{figure}[t]
  \centering
  \includegraphics[width=\columnwidth]{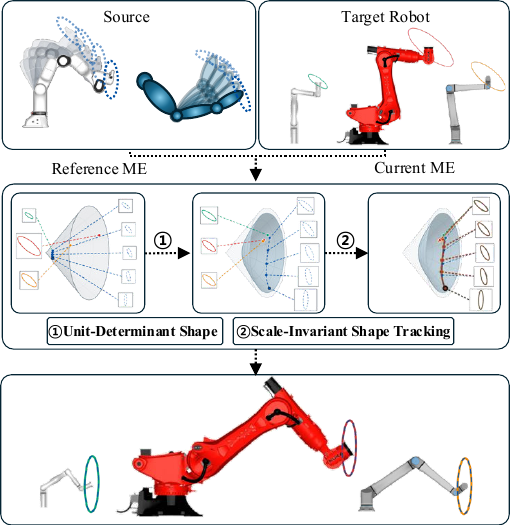}
  \caption{Manipulability ellipsoid (ME) shape transfer from robot or human motion to heterogeneous robots.
\textcircled{\scriptsize 1}~Reference and current manipulability matrices are normalized
to unit-determinant shapes, preserving ME orientation and
all semi-axis length ratios.
\textcircled{\scriptsize 2}~The Shape method follows the reference
shape trajectory without requiring absolute-scale matching.
Blue dashed ellipses denote reference MEs; green, red, and orange solid ellipses denote the MEs of Gen3, KR500, and UR20, respectively.}
\vspace{-0.6cm}
  \label{fig:shape-transfer-overview}
\end{figure}

\begin{figure*}[t]
  \centering
  \includegraphics[width=\textwidth]{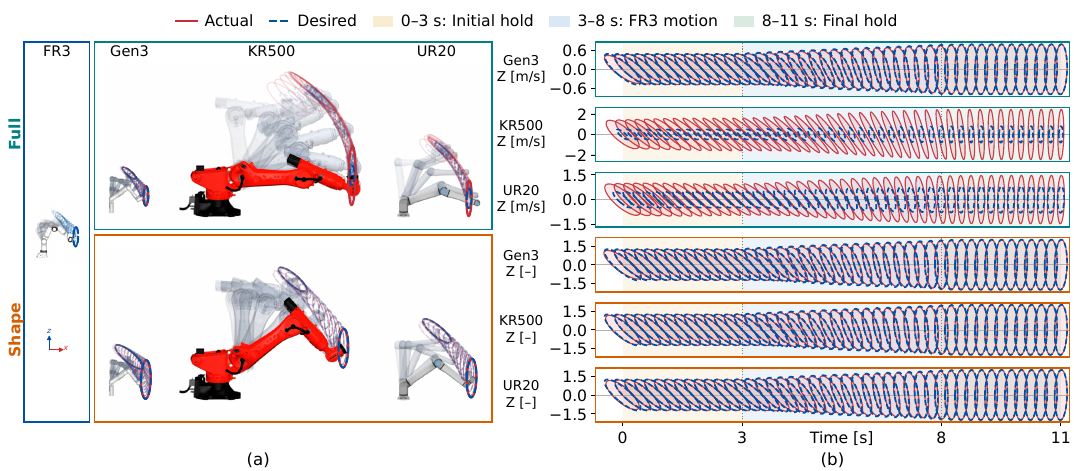}
  \caption{Manipulability transfer from FR3 to Gen3, KR500, and UR20.
(a) Robot motions: FR3 source (blue frame) and followers using
the Full (teal) and Shape (orange) methods.
(b) Actual (red solid) and desired (blue dashed) XZ ellipses
centered at the corresponding times, displayed as $\bm M$
for the Full method and $\widehat{\bm M}$ for the Shape method.}
\vspace{-0.5cm}
  \label{fig:heterogeneous-manipulability-transfer}
\end{figure*}
When transferring MEs between different kinematic systems,
identical orientations and semi-axis length ratios can coexist
with different absolute scales.
For example, consider revolute-joint manipulators with the
same joint arrangement and link-length ratios but a uniform
scaling of all linear dimensions.
At corresponding joint configurations and under identical
joint-velocity norm constraints, their translational velocity
MEs have the same orientation and semi-axis length ratios
but different sizes.
More general kinematic differences can affect both directional
structure and scale.
Reproducing the orientation and semi-axis length ratios should
therefore be distinguished from matching the absolute scale.

To connect different manipulability domains, Reithmeir et al.
proposed alignment using parallel transport and manifold-aware
iterative closest point (ICP) \cite{reithmeir2022human}.
More recently, Gong et al. first estimated a single global scaling factor for the demonstration set by numerical optimization, then applied parallel transport with manipulability feasibility checks at future EE poses, and constrained tracking via quadratic programming (QP) \cite{gong2025manipulability}.
Both approaches retain the absolute scale in the reference and adapt it to the robot before tracking. 
Here, we focus on transferring ME \emph{shape}, defined as
its orientation in a common task frame and all semi-axis
length ratios, without requiring absolute-scale matching.
This calls for a tracking objective that excludes uniform-scale
differences between the current and reference MEs.

We propose a scale-invariant manipulability shape-tracking
approach (the \emph{Shape} method) under the affine-invariant
Riemannian metric (AIRM).
Our contributions are threefold.
(i) We formulate scale-invariant manipulability transfer
by treating SPD matrices that differ only in absolute scale
as equivalent and using their unit-determinant representatives
(Fig.~\ref{fig:shape-transfer-overview}).
This normalization also preserves velocity--force duality
under the stated norm conventions.
(ii) We derive the differential of the unit-determinant
shape representative and an orthonormal coordinate
representation of the AIRM tangent tracking residual.
(iii) We show that, with the same scalar feedback gain,
the AIRM-based full-matrix tracking cost decomposes exactly
into the shape cost and a scalar log-scale rate-tracking term.
The proposed objective therefore excludes the reference scale
that existing methods track directly \cite{jaquier2021geometry}
or first align to the robot
\cite{reithmeir2022human,gong2025manipulability}.

The experiments further show what a constant scale factor
can and cannot do in place of the omitted scale term.
A multiplier fitted for a given robot during motion brings
the Full method's shape tracking close to that of the Shape
method in that phase.
It still leaves endpoint shape errors on the larger arms
and is not necessarily suitable for another robot.
Integrated with position and EE direction objectives in a
constrained joint-velocity QP, the Shape method transfers
manipulability shape between robots of different kinematic
structures and from a recorded human reach, without selecting
a target ME scale.

\section{Background}
\label{sec:background}

\subsection{Manipulability Ellipsoids}

Let $\bm{q}\in\mathbb R^n$ denote the active joint coordinates
and $\dot{\bm{x}}\in\mathbb R^D$ the Cartesian task velocity,
related by $\dot{\bm{x}}=\bm{J}(\bm{q})\dot{\bm{q}}$.
We assume that $\bm{J}\in\mathbb R^{D\times n}$ has full
row rank at the configurations considered.
The velocity manipulability matrix is
\begin{equation}
\bm{M}=\bm{J}\bm{J}^{\top}\in\mathcal S_{++}^{D},
\label{eq:manipulability_matrix}
\end{equation}
where $\mathcal S_{++}^{D}$ is the set of $D\times D$
symmetric positive-definite (SPD) matrices.
The image of the Euclidean joint-velocity unit ball
$\|\dot{\bm{q}}\|_2\leq1$ is the manipulability ellipsoid
\cite{yoshikawa1985manipulability,jaquier2018geometry}
\begin{equation}
\mathcal E(\bm{M})
=
\left\{
\dot{\bm{x}}\in\mathbb R^D \,|\, 
\dot{\bm{x}}^{\top}\bm{M}^{-1}\dot{\bm{x}}\leq1
\right\}.
\label{eq:velocity_me}
\end{equation}
Its principal directions are the eigenvectors of $\bm{M}$,
and its semi-axis lengths are $\sqrt{\lambda_i(\bm{M})}$,
where $\lambda_i(\bm{M})$ denotes the $i$th eigenvalue of
$\bm{M}$.

Dually, the static relation
$\bm{\tau}=\bm{J}^{\top}\bm{f}$ maps task-space forces
to joint torques.
Under the Euclidean torque bound $\|\bm{\tau}\|_2\leq1$,
admissible forces satisfy $\bm{f}^{\top}\bm{M}\bm{f}\leq1$.
Thus, the force manipulability matrix is
$\bm{M}_f=\bm{M}^{-1}$, giving the same principal axes
with reciprocal corresponding semi-axis lengths
\cite{jaquier2018geometry}.

\begin{figure}[t]
  \centering
  \includegraphics[width=\columnwidth]{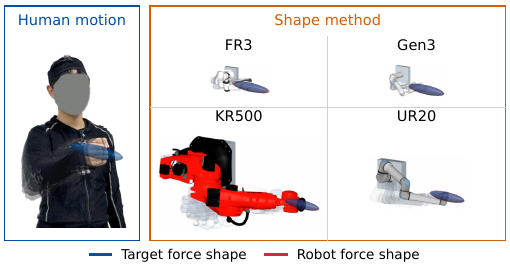}
  \vspace{-0.8cm}
  \caption{Human-to-robot transfer of wrist position and manipulability
shape from a measured right-arm reach.
Four heterogeneous manipulators use the Shape method to track the
transferred position and velocity manipulability shape with
a fixed forward end-effector direction objective.
Ellipsoids show the final human target (blue) and robot (red)
dual force shapes.}
\vspace{-0.5cm}
  \label{fig:human-to-robot-shape-motion}
\end{figure}

\subsection{Riemannian Geometry of SPD Matrices}

We equip $\mathcal S_{++}^{D}$ with the affine-invariant
Riemannian metric (AIRM)
\cite{moakher2005differential,jaquier2021geometry,pennec2006riemannian}.
At $\bm{P}\in\mathcal S_{++}^{D}$, the tangent space is
identified with $\operatorname{Sym}(D)$, the space of
real symmetric matrices.
For tangent vectors $\bm{U},\bm{V}$, the inner product is
$\langle\bm{U},\bm{V}\rangle_{\bm{P}}
=\operatorname{tr}(\bm{P}^{-1}\bm{U}\bm{P}^{-1}\bm{V})$,
and its induced norm satisfies
\begin{equation}
\|\bm{U}\|_{\bm{P}}^2
=
\left\|
\bm{P}^{-1/2}\bm{U}\bm{P}^{-1/2}
\right\|_F^2,
\label{eq:airm_tangent_norm}
\end{equation}
where $\|\cdot\|_F$ is the Frobenius norm and
$\bm{P}^{1/2}$ denotes the SPD square root.

The mutually inverse Riemannian exponential and logarithm
maps are
\begin{align}
\operatorname{Exp}_{\bm{P}}(\bm{U})
&=
\bm{P}^{1/2}
\exp\!\left(\bm{P}^{-1/2}\bm{U}\bm{P}^{-1/2}\right)
\bm{P}^{1/2},
\label{eq:airm_exp}\\
\operatorname{Log}_{\bm{P}}(\bm{Q})
&=
\bm{P}^{1/2}
\log\!\left(\bm{P}^{-1/2}\bm{Q}\bm{P}^{-1/2}\right)
\bm{P}^{1/2},
\label{eq:airm_log}
\end{align}
where $\exp$ and $\log$ denote the matrix exponential
and principal matrix logarithm.
The corresponding geodesic distance is
\begin{equation}
d_{\mathrm{AI}}(\bm{P},\bm{Q})
=
\left\|
\log\!\left(\bm{P}^{-1/2}\bm{Q}\bm{P}^{-1/2}\right)
\right\|_F,
\label{eq:airm_distance}
\end{equation}
with
$\|\operatorname{Log}_{\bm{P}}(\bm{Q})\|_{\bm{P}}
=d_{\mathrm{AI}}(\bm{P},\bm{Q})$.

\section{Scale-Invariant Manipulability Shape Tracking}
\label{sec:method}

We refer to the full-matrix tracking baseline as the \emph{Full}
method and the proposed scale-invariant shape-tracking approach
as the \emph{Shape} method.
The Full method tracks the complete reference ME, including scale
and shape, whereas the Shape method tracks its orientation and all
semi-axis length ratios without requiring absolute-scale matching.

\subsection{Manipulability Differential and Full-Matrix Tracking}
\label{sec:manipulability_differential}

Following the tensor notation and matrix differentiation
rules in \cite{jaquier2021geometry}, define the
third-order manipulability Jacobian
$\bm{\mathcal{J}}=\partial\bm{M}/\partial\bm{q}
\in\mathbb R^{D\times D\times n}$.
The product and chain rules give
% \begin{align}
% \bm{\mathcal{J}}
% &=\frac{\partial\bm{J}}{\partial\bm{q}}\times_2\bm{J}
% +\frac{\partial\bm{J}^{\top}}{\partial\bm{q}}\times_1\bm{J},
% \label{eq:manipulability_jacobian}\\
% \dot{\bm{M}}
% &=\bm{\mathcal{J}}\times_3\dot{\bm{q}}^{\top}.
% \label{eq:manipulability_rate}
% \end{align}
\begin{equation}
\bm{\mathcal{J}}
=\frac{\partial\bm{J}}{\partial\bm{q}}\times_2\bm{J}
+\frac{\partial\bm{J}^{\top}}{\partial\bm{q}}\times_1\bm{J},
\quad
\dot{\bm{M}}=\bm{\mathcal{J}}\times_3\dot{\bm{q}}^{\top}.
\label{eq:manipulability_jacobian}
\end{equation}
Here, $\times_k$ denotes the mode-$k$ product, and
$\bm{\mathcal{J}}_{:,:,j}=\partial\bm{M}/\partial q_j$
is the $j$th matrix slice.
The mode-3 contraction is identified with a $D\times D$ matrix.
All derivatives are evaluated at the current configuration.

Let $\bm{M}_c=\bm{M}(\bm{q})$ and
$\bm{M}_d\in\mathcal S_{++}^{D}$ denote the current and
reference matrices.
For scalar gain $k_M>0$, define the rate residual and cost of the Full method:
\begin{align}
\bm{\mathcal{R}}_{\mathrm{full}}(\dot{\bm{q}})
&=\bm{\mathcal{J}}\times_3\dot{\bm{q}}^{\top}
-k_M\operatorname{Log}_{\bm{M}_c}(\bm{M}_d),
\label{eq:full_tracking_residual}\\
C_{\mathrm{full}}(\dot{\bm{q}})
&=\frac12\left\|\bm{\mathcal{R}}_{\mathrm{full}}
(\dot{\bm{q}})\right\|_{\bm{M}_c}^2.
\label{eq:full_tracking_cost}
\end{align}
The residual compares the robot-induced rate with the
logarithmic feedback command in
$T_{\bm{M}_c}\mathcal S_{++}^{D}$.
The Full method adopts the logarithmic feedback of
\cite{jaquier2021geometry} with a scalar gain,
but measures the rate residual in the AIRM tangent norm
instead of the Euclidean norm used in the original
vectorized least-squares formulation.

\subsection{Unit-Determinant Shape Representation}
\label{sec:scale_shape_decomposition}

For $\bm{M}\in\mathcal S_{++}^{D}$, the determinant-based
decomposition \cite{moakher2005differential} defines the
matrix scale $\rho$ and unit-determinant shape
representative $\widehat{\bm{M}}$ as
\begin{equation}
\rho=\rho(\bm{M})=(\det\bm{M})^{1/D}>0,
\qquad \widehat{\bm{M}}=\rho^{-1}\bm{M}.
\label{eq:normalized_manipulability}
\end{equation}
Thus, $\bm{M}=\rho\widehat{\bm{M}}$ and
$\widehat{\bm{M}}\in\mathcal{SP}(D)
=\{\bm{P}\in\mathcal S_{++}^{D} \,|\,\det\bm{P}=1\}$.
For manipulability matrices, hats denote determinant normalization.
Normalization divides every semi-axis length by $\sqrt{\rho}$,
preserving the principal directions and all semi-axis ratios.
For any $a>0$, $\rho(a\bm{M})=a\rho(\bm{M})$, so
$a\bm{M}/\rho(a\bm{M})=\widehat{\bm{M}}$.
The Shape method therefore compares $\widehat{\bm{M}}_c$ and
$\widehat{\bm{M}}_d$ without matching their original scales.
The achieved scale remains configuration-dependent and need
not remain constant.

\subsection{Scale-Invariant Shape Tracking}
\label{sec:shape_differential}

Let
$\widehat{\bm{\mathcal{J}}}
=\partial\widehat{\bm{M}}/\partial\bm{q}
\in\mathbb R^{D\times D\times n}$
denote the derivative tensor of the unit-determinant
shape matrix $\widehat{\bm{M}}$.
Jacobi's formula and the chain rule give
$\partial\log\det\bm{M}/\partial q_j
=\operatorname{tr}(\bm{M}^{-1}\bm{\mathcal{J}}_{:,:,j})$.
Differentiating \eqref{eq:normalized_manipulability} yields
\begin{equation}
\widehat{\bm{\mathcal{J}}}_{:,:,j}
=
\rho^{-1}
\left[
\bm{\mathcal{J}}_{:,:,j}
-
\frac{
\operatorname{tr}(\bm{M}^{-1}\bm{\mathcal{J}}_{:,:,j})
}{D}\bm{M}
\right].
\label{eq:normalized_manipulability_jacobian}
\end{equation}
The second term accounts for the configuration-dependent
scale, and the shape rate satisfies
$\dot{\widehat{\bm{M}}}
=\widehat{\bm{\mathcal{J}}}\times_3\dot{\bm{q}}^{\top}$.

At $\bm{P}\in\mathcal{SP}(D)$, the tangent space is
$T_{\bm{P}}\mathcal{SP}(D)
=\{\bm{U}\in\operatorname{Sym}(D)\,|\,
\operatorname{tr}(\bm{P}^{-1}\bm{U})=0\}$
\cite{dolcetti2019differential}.
The trace condition follows by differentiating
$\log\det\bm{P}(t)=0$ along a curve in $\mathcal{SP}(D)$.
With $m=D(D+1)/2$, this space has dimension $m-1$.
The transformation
$\bm{X}=\bm{P}^{-1/2}\bm{U}\bm{P}^{-1/2}$
maps it to $\operatorname{Sym}_0(D)$, the space of real
symmetric traceless matrices.

Choose a fixed Frobenius-orthonormal basis
$\{\bm{B}_{\alpha}\}_{\alpha=1}^{m-1}$ of
$\operatorname{Sym}_0(D)$, using the real symmetric
off-diagonal and diagonal generalized Gell--Mann matrices
normalized to unit Frobenius norm \cite{bertlmann2008bloch}.
For $\bm{X}\in\operatorname{Sym}_0(D)$, define
$[\operatorname{vec}_0(\bm{X})]_{\alpha}
=\operatorname{tr}(\bm{B}_{\alpha}\bm{X})$,
$\alpha=1,\ldots,m-1$, with the explicit basis and
2D/3D coordinate expressions given in Appendix~A.
For $\bm{U}\in T_{\bm{P}}\mathcal{SP}(D)$, define
\begin{equation}
\mathcal C_{\bm{P}}(\bm{U})
=\operatorname{vec}_0
\left(\bm{P}^{-1/2}\bm{U}\bm{P}^{-1/2}\right).
\label{eq:shape_tangent_coordinates}
\end{equation}
At fixed $\bm{P}$, this map is linear.
Basis orthonormality and \eqref{eq:airm_tangent_norm} give
$\|\mathcal C_{\bm{P}}(\bm{U})\|_2^2
=\|\bm{U}\|_{\bm{P}}^2$.

At the current shape $\widehat{\bm{M}}_c$, define the
logarithmic error coordinates $\bm{e}_s\in\mathbb R^{m-1}$ as
\begin{equation}
\bm{e}_s
=
\operatorname{vec}_0
\left[
\log\!\left(
\widehat{\bm{M}}_c^{-1/2}
\widehat{\bm{M}}_d
\widehat{\bm{M}}_c^{-1/2}
\right)
\right].
\label{eq:shape_error}
\end{equation}
The relative matrix inside the logarithm is SPD with unit
determinant, so its logarithm is symmetric and traceless.
Thus,
$\bm{e}_s
=\mathcal C_{\widehat{\bm{M}}_c}
(\operatorname{Log}_{\widehat{\bm{M}}_c}
(\widehat{\bm{M}}_d))$.

The coordinate Jacobian
$\bm{J}_M\in\mathbb R^{(m-1)\times n}$
is defined columnwise by
\begin{equation}
(\bm{J}_M)_{:,j}
=
\mathcal C_{\widehat{\bm{M}}_c}
\left(\widehat{\bm{\mathcal{J}}}_{:,:,j}\right),
\label{eq:shape_coordinate_jacobian}
\end{equation}
with the derivative tensor evaluated at the current
joint configuration.
By linearity,
$\bm{J}_M\dot{\bm{q}}
=\mathcal C_{\widehat{\bm{M}}_c}
(\dot{\widehat{\bm{M}}}_c)$;
this represents the current shape rate, not
$\dot{\bm{e}}_s$.

The Shape method uses the same scalar gain $k_M$ as the Full
method, with the tracking cost
\begin{equation}
C_{\mathrm{shape}}(\dot{\bm{q}})
=
\frac{1}{2}
\left\|\bm{J}_M\dot{\bm{q}}-k_M\bm{e}_s\right\|_2^2.
\label{eq:shape_tracking_cost}
\end{equation}
The shape rate and logarithmic feedback belong to
$T_{\widehat{\bm{M}}_c}\mathcal{SP}(D)$.
Therefore, the norm-preserving property of
$\mathcal C_{\widehat{\bm{M}}_c}$ gives
\begin{equation}
C_{\mathrm{shape}}(\dot{\bm{q}})
=
\frac{1}{2}
\left\|
\widehat{\bm{\mathcal{J}}}\times_3\dot{\bm{q}}^{\top}
-
k_M\operatorname{Log}_{\widehat{\bm{M}}_c}
(\widehat{\bm{M}}_d)
\right\|_{\widehat{\bm{M}}_c}^{2}.
\label{eq:shape_tracking_cost_airm}
\end{equation}
Appendix~A establishes this equivalence.
The shape tangent residual is represented exactly by
$m-1$ coordinates: two for $D=2$ and five for $D=3$.
This reduction concerns the tangent representation; it does not guarantee that $\bm{J}_M$ has full row rank or that the reference shape is reachable.
Exact tracking may be limited by the robot's kinematics, joint constraints, or competing task objectives.

\subsection{Geometric Properties}
\label{sec:geometric_properties}

\emph{Scale invariance.}
Let $\rho_c=\rho(\bm{M}_c)$ and
$\rho_d=\rho(\bm{M}_d)$ denote the current and reference
matrix scales, and let
$e_\rho=\log(\rho_d/\rho_c)$.
Define the relative shape matrix
$\bm{R}_s
=\widehat{\bm{M}}_c^{-1/2}
\widehat{\bm{M}}_d
\widehat{\bm{M}}_c^{-1/2}$.
The scale--shape decomposition gives
$\bm{M}_c^{-1/2}\bm{M}_d\bm{M}_c^{-1/2}
=(\rho_d/\rho_c)\bm{R}_s$.
Since $\rho_d/\rho_c$ is a positive scalar,
\begin{equation}
\log\!\left(
\bm{M}_c^{-1/2}\bm{M}_d\bm{M}_c^{-1/2}
\right)
=
e_\rho\bm{I}+\log\bm{R}_s,
\label{eq:relative_log_decomposition}
\end{equation}
where $\bm{I}$ is the $D\times D$ identity matrix.
Since $\det\bm{R}_s=1$,
$\operatorname{tr}(\log\bm{R}_s)=0$.
The two components are therefore Frobenius-orthogonal,
and taking squared norms with $\|\bm{I}\|_F^2=D$ yields
\cite{moakher2005differential}
\begin{equation}
d_{\mathrm{AI}}^2(\bm{M}_c,\bm{M}_d)
=
d_{\mathrm{AI}}^2
(\widehat{\bm{M}}_c,\widehat{\bm{M}}_d)
+
D e_\rho^2.
\label{eq:scale_shape_distance}
\end{equation}
The squared shape-distance term equals $\|\bm{e}_s\|_2^2$.
For any $a>0$, replacing $\bm{M}_d$ by $a\bm{M}_d$
leaves $\widehat{\bm{M}}_d$ unchanged.
Thus, at a fixed current configuration, both $\bm{e}_s$
and $C_{\mathrm{shape}}(\dot{\bm{q}})$ are invariant to target rescaling.

\emph{Decomposition of the tracking cost.}
Consider both costs at the same current configuration,
reference, and scalar gain $k_M$.
Since $\log\rho=(\log\det\bm{M})/D$, Jacobi's formula
gives the logarithmic scale Jacobian
$\bm{J}_\rho\in\mathbb R^{1\times n}$ at the current
configuration:
\begin{equation}
(\bm{J}_\rho)_{1,j}
=\frac{\partial\log\rho}{\partial q_j}
=\frac{1}{D}\operatorname{tr}
\left(\bm{M}_c^{-1}\bm{\mathcal{J}}_{:,:,j}\right).
\label{eq:log_scale_jacobian}
\end{equation}
Differentiating $\bm{M}_c=\rho_c\widehat{\bm{M}}_c$
and using $\bm{J}_\rho\dot{\bm{q}}=\dot{\rho}_c/\rho_c$
together with \eqref{eq:relative_log_decomposition},
we write the transformed residual
$\bm{M}_c^{-1/2}\bm{\mathcal{R}}_{\mathrm{full}}(\dot{\bm{q}})
\bm{M}_c^{-1/2}$
as the sum of the traceless shape residual
$\widehat{\bm{M}}_c^{-1/2}\dot{\widehat{\bm{M}}}_c
\widehat{\bm{M}}_c^{-1/2}-k_M\log\bm{R}_s$
and the scale residual
$(\bm{J}_\rho\dot{\bm{q}}-k_Me_\rho)\bm{I}$.
The shape residual has squared Frobenius norm
$2C_{\mathrm{shape}}(\dot{\bm{q}})$ by \eqref{eq:shape_tracking_cost_airm}.
Frobenius orthogonality, with $\|\bm{I}\|_F^2=D$, then gives
\begin{equation}
C_{\mathrm{full}}(\dot{\bm{q}})
=C_{\mathrm{shape}}(\dot{\bm{q}})
+\frac{D}{2}
\left(\bm{J}_\rho\dot{\bm{q}}-k_Me_\rho\right)^2.
\label{eq:full_shape_cost_decomposition}
\end{equation}
The Shape method therefore omits precisely the scalar log-scale
rate-tracking term from its ME cost.
This does not imply independent control of scale and
shape through joint motion.

\emph{Preservation of velocity--force duality.}
Under the norm conventions in Section~\ref{sec:background},
$\bm{M}_f=\bm{M}^{-1}$ and
$\rho_f=(\det\bm{M}_f)^{1/D}=\rho^{-1}$.
Normalizing the force matrix gives
\begin{equation}
\widehat{\bm{M}}_f
=\rho_f^{-1}\bm{M}_f
=\rho\bm{M}^{-1}
=\widehat{\bm{M}}^{-1}.
\label{eq:normalized_velocity_force_duality}
\end{equation}
The unit-determinant velocity and force shapes therefore
share principal axes with reciprocal corresponding
semi-axis lengths.

\subsection{Integration with Task-Space Control}
\label{sec:joint_velocity_qp}

For each task $\ell=1,\ldots,N_t$, let
$\bm J_\ell\in\mathbb R^{d_\ell\times n}$ and
$\bm v_\ell^{\mathrm{cmd}}\in\mathbb R^{d_\ell}$
denote its Jacobian and velocity command, expressed in
the same task-velocity coordinates.
With weights $w_\ell\geq0$, the combined task-space cost is
\begin{equation}
C_{\mathrm{task}}(\dot{\bm q})
=\sum_{\ell=1}^{N_t}\frac{w_\ell}{2}
\left\|\bm J_\ell\dot{\bm q}
-\bm v_\ell^{\mathrm{cmd}}\right\|_2^2.
\label{eq:task_space_cost}
\end{equation}

The joint-velocity command is obtained from
\begin{equation}
\begin{aligned}
\dot{\bm q}^{*}
=\arg\min_{\dot{\bm q}\in\mathcal U(\bm q)}C_{\mathrm{task}}(\dot{\bm q})+w_M C_M(\dot{\bm q})+\frac{\eta^2}{2}\|\dot{\bm q}\|_2^2,
\end{aligned}
\label{eq:joint_velocity_qp}
\end{equation}
where $C_M=C_{\mathrm{shape}}$ for the Shape method and
$C_M=C_{\mathrm{full}}$ for the Full method.
Here, $w_M\geq0$ is the ME-tracking weight,
$\eta>0$ is the regularization parameter, and
$\mathcal U(\bm q)$ imposes joint-velocity limits and
one-step joint-position bounds.
Task-space and ME residuals are minimized jointly as
weighted soft objectives; for ME-only tracking,
$C_{\mathrm{task}}=0$.
Both methods use matched weights, feedback gains,
regularization, and constraints.
The task choices and parameter values are specified
in the experiments.

\begin{table}[t]
  \makeatletter
  \long\def\@makecaption#1#2{%
    {\centering\footnotesize #1.\enspace{\scshape #2}\par}%
    \vskip 5pt}
  \makeatother
  \caption{Tracking errors}
  \label{tab:transfer-errors}
  \centering
  \footnotesize
  \setlength{\tabcolsep}{2pt}
  \renewcommand{\arraystretch}{1.15}
  \begin{tabular*}{\columnwidth}{@{\extracolsep{\fill}}cl|ccc|ccc@{}}
    \hline
    \multicolumn{8}{c}{Phase means} \\
    \hline
    &
    & \multicolumn{3}{c|}{Full} & \multicolumn{3}{c}{Shape} \\
    \hline
    \rule{0pt}{3ex}Robot & Phase [s]
    & $\overline d_{\mathrm{AI}}$ & $\overline d_s$ & $\overline d_\rho$
    & $\overline d_{\mathrm{AI}}$ & $\overline d_s$ & $\overline d_\rho$ \\
    \hline
    Gen3 & $[0,3)$ & 0.0348 & 0.0344 & 0.0055 & 0.0651 & 0.0344 & 0.0421 \\
     & $[3,8)$ & 0.2210 & 0.2210 & 0.0033 & 0.2234 & 0.2210 & 0.0176 \\
     & $[8,11]$ & 0.0409 & 0.0408 & 0.0021 & 0.0471 & 0.0408 & 0.0101 \\
    \hline
    KR500 & $[0,3)$ & 2.7018 & 0.8115 & 2.5136 & 3.0878 & 0.0243 & 3.0873 \\
     & $[3,8)$ & 2.7768 & 1.7640 & 2.1402 & 3.1048 & 0.2210 & 3.0964 \\
     & $[8,11]$ & 2.4271 & 1.1999 & 2.1044 & 3.0303 & 0.0408 & 3.0295 \\
    \hline
    UR20 & $[0,3)$ & 1.6226 & 0.5865 & 1.4650 & 2.0148 & 0.0292 & 2.0139 \\
     & $[3,8)$ & 1.5223 & 0.9075 & 1.2215 & 2.0291 & 0.2210 & 2.0162 \\
     & $[8,11]$ & 1.2444 & 0.7499 & 0.9927 & 1.9362 & 0.0408 & 1.9349 \\
    \hline
  \end{tabular*}
\end{table}

\begin{table}[t]
  \makeatletter
  \long\def\@makecaption#1#2{%
    {\centering\footnotesize #1.\enspace{\scshape #2}\par}%
    \vskip 5pt}
  \makeatother
  \caption{Target-scale sensitivity of the Full method}
  \label{tab:scale-errors}
  \centering
  \footnotesize
  \setlength{\tabcolsep}{2pt}
  \renewcommand{\arraystretch}{1.15}
  % Center the multiplier in its column; stretch only the metric-column gaps.
  \begin{tabular*}{\columnwidth}{@{}cc|@{\extracolsep{\fill}}ccc@{}}
    \hline
    \multicolumn{5}{c}{Endpoint ($t=11$~s)} \\
    \hline
    Follower & \makebox[.27\columnwidth][c]{Multiplier $c$} & $|r-r_d|$ & $\theta\,[\mathrm{deg}]$ & $d_s$ \\
    \hline
    Gen3 & 1.189 & 0.0002 & 0.0008 & 0.0001 \\
         & 11.31 & 1.0687 & 0.0005 & 0.4352 \\
         & 5.657 & 0.9562 & 0.0005 & 0.3826 \\
    \hline
    KR500 & 1.189 & 4.5546 & 0.3019 & 1.0694 \\
          & 11.31 & 0.1905 & 0.0004 & 0.0684 \\
          & 5.657 & 0.4957 & 0.0004 & 0.1639 \\
    \hline
    UR20 & 1.189 & 2.2041 & 0.0228 & 0.6164 \\
         & 11.31 & 0.7781 & 0.0007 & 0.3030 \\
         & 5.657 & 0.3054 & 0.0006 & 0.1113 \\
    \hline
  \end{tabular*}
  \vspace{-0.5cm}
\end{table}

\section{Experiments}
\label{sec:experiments}

\subsection{Experimental Setup}
\label{sec:experimental_setup}

Experiments were conducted as kinematic simulations using MuJoCo 3.5.0.
At each $2$~ms simulation step ($500$~Hz), the joint-velocity QP
was solved and the joint configuration was updated by forward Euler
integration, with joint speeds bounded by $0.6$~rad/s.
The joint-velocity QPs for manipulability transfer were solved using
the Python package \texttt{quadprog} 0.1.13.
For each robot, the Full and Shape methods used the same initial
configuration and source motion, with matched control settings.
Constant ME target-scale multipliers are applied to the Full method
only in Section~\ref{sec:target_scale_experiment}.
Initial holds provide time to reduce the initial manipulability
mismatch before motion begins, while final holds allow residual
tracking errors to be assessed after the source position reference stops.

To compare different sizes, joint counts, and link proportions, we selected
the seven-joint Franka Research 3 (FR3) from Franka Robotics and Gen3 from
Kinova, and the six-joint KR~500 R2800-2 (KR500) from KUKA and UR20 from
Universal Robots.
The reference arm length $L_r$ was measured in a planar configuration
with joints not used for XZ motion fixed: it is the vertical distance from
the first active joint to the tool center point (TCP)
at maximum extension along $+Z$.
For FR3, Gen3, KR500, and UR20, $L_r$ is $0.8579$, $0.9026$, $2.5863$,
and $1.7500$~m, respectively.
Despite their similar lengths and identical joint counts, FR3 and Gen3
have different ratios of the first two effective XZ link lengths,
$\ell_1/\ell_2=0.832$ and $1.338$, respectively.

\begin{figure}[t]
  \centering
  \includegraphics[width=\columnwidth]{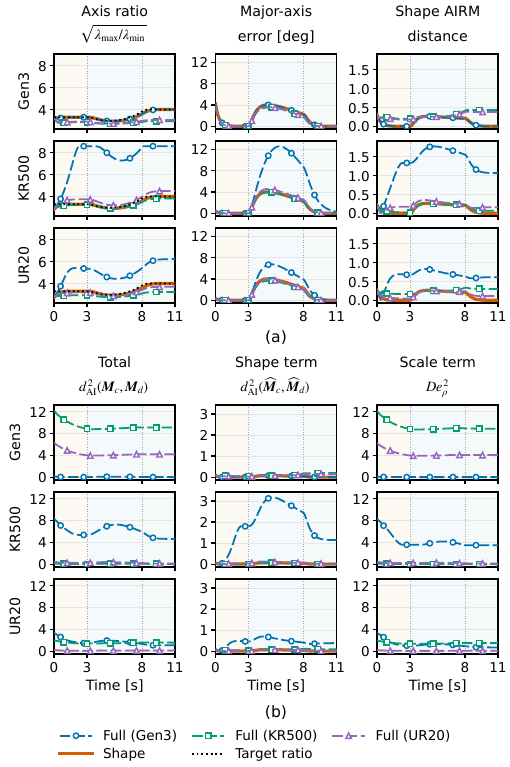}
  \caption{Target-scale sensitivity with
$\bm M_d^{(c)}=c\bm M_{\mathrm{FR3}}$.
Full (Gen3), Full (KR500), and Full (UR20) use
$c=1.189,\,11.31,\,5.657$, respectively.
Each multiplier is applied to all followers.
(a) Axis ratio, major-axis alignment error, and shape AIRM distance.
(b) Squared total AIRM distance for the Full method, decomposed
into shape and scale terms; orange shows the squared shape
AIRM distance for the Shape method.}
\vspace{-0.5cm}
  \label{fig:target-scale-sensitivity}
\end{figure}

\begin{figure*}[t]
  \centering
  \includegraphics[width=\textwidth]{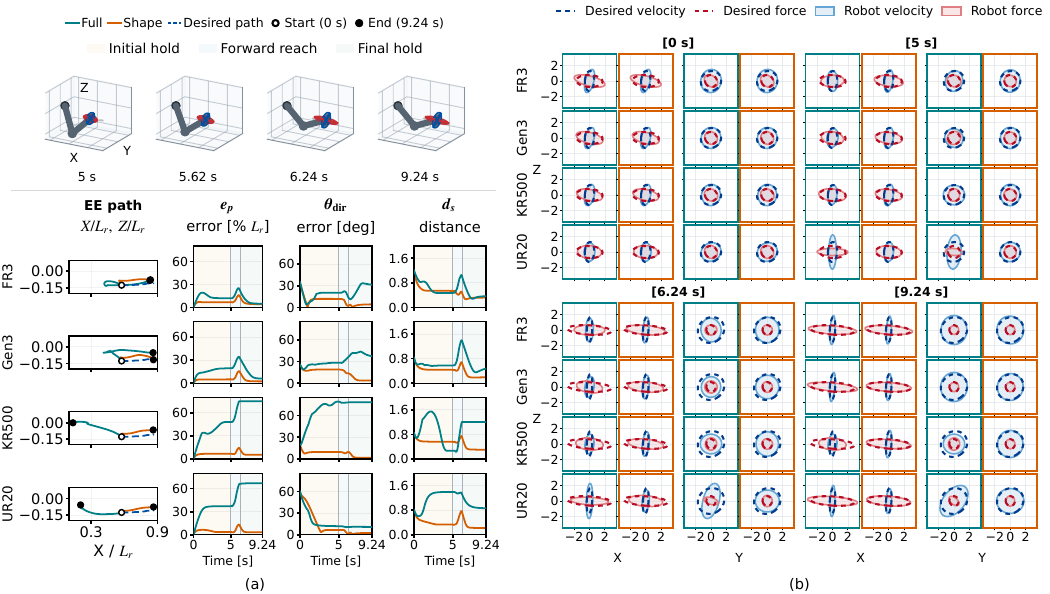}
  \caption{Human-to-robot ME transfer with concurrent EE position and direction tasks.
Full (teal) tracks the original human ME; Shape (orange) tracks its unit-determinant shape representative.
(a) Human snapshots, EE paths, and position error $e_p$, EE direction error
$\theta_{\mathrm{dir}}$, and shape AIRM distance $d_s$ over $0$--$9.24$~s. EE paths are shown relative to $\bm o_r$ and normalized by $L_r$.
(b) XZ and YZ shape projections at $0$ and $5$~s (top), and $6.24$ and $9.24$~s (bottom).
At each time, the left and right pairs show XZ and YZ projections, respectively.
Teal and orange frames identify the Full and Shape methods, respectively.
Blue and red contours denote normalized velocity and dual force shapes;
dashed contours denote human targets and filled contours robot results.}
\vspace{-0.5cm}
  \label{fig:human-transfer-full-shape-xz-yz}
\end{figure*}

\subsection{Robot-to-Robot Manipulability Transfer}
\label{sec:robot_transfer_experiment}

The first experiment transfers the same time-varying FR3 ME,
$\bm M_d(t)$, to Gen3, KR500, and UR20, comparing the Full method,
which tracks the complete matrices, with the Shape method,
which tracks their unit-determinant shape representatives.
Each robot uses three active joints for XZ motion, and the
followers track only manipulability, without position or
direction objectives. For ME-only tracking, we used $w_M=1$, $k_M=3$, and
$\eta=0.002$, without target-rate feedforward.
The FR3 TCP reference specifies a straight motion forward ($+X$)
and downward ($-Z$) in the world XZ plane.
It is held at $(x,z)=(0.300,0.847)$~m over $0$--$3$~s,
moves at constant speed to $(0.650,0.417)$~m over $3$--$8$~s,
and is then held until $11$~s.
The ME reference is computed from the realized FR3 configuration
and can therefore continue changing during the final position hold.
Fig.~\ref{fig:heterogeneous-manipulability-transfer} shows
(a) the robot motions and (b) the actual and target MEs over time.

At each time sample, we take the square root of each term in
\eqref{eq:scale_shape_distance} to obtain the total, shape,
and scale distances.
% Table~\ref{tab:transfer-errors} reports their phase means
% $\overline d_{\mathrm{AI}}$, $\overline d_s$, and
% $\overline d_\rho$, where the overbar denotes the arithmetic
% mean of the distances within each phase.
Table~\ref{tab:transfer-errors} reports their phase means
$\overline d_{\mathrm{AI}}$, $\overline d_s$, and
$\overline d_\rho$.
An overbar denotes the arithmetic mean of the indicated
quantity over the specified phase.
Here, $d_s=d_{\mathrm{AI}}(\widehat{\bm M}_c,\widehat{\bm M}_d)$
and $d_\rho=\sqrt{2}|e_\rho|$ for $D=2$.
Both methods are evaluated against the same original target
$\bm M_d$ over $[0,3)$, $[3,8)$, and $[8,11]$~s.

Gen3 has nearly identical mean shape errors with both methods,
whereas KR500 and UR20 have smaller errors with the Shape method
in every phase.
During the final hold, the Full method yields
$\overline d_s=1.1999$ for KR500 and $0.7499$ for UR20,
while the Shape method yields approximately $0.0408$
for all three followers.
At the endpoint, the Shape method achieves
$d_s(11)\approx9.30\times10^{-5}$ for all three followers,
whereas the Full method retains $d_s(11)=1.0696$ for KR500
and $0.7576$ for UR20.

Conversely, the Shape method has a larger mean total AIRM
distance for every follower in every phase; KR500 and UR20
also have larger mean scale distances with the Shape method
throughout.
The total distance includes absolute-scale differences,
which the Shape method does not penalize.
Total and shape distances should therefore be distinguished
when interpreting tracking performance.
These results show that, for the tested robots with different
kinematic structures, the Shape method reproduces the target
ME orientation and semi-axis length ratios with small endpoint
residuals without requiring absolute-scale matching.

\subsection{Target-Scale Sensitivity}
\label{sec:target_scale_experiment}

The second experiment asks whether a constant target scale can take the place of the scale term that the proposed objective omits.
It evaluates how robot-specific target scaling improves shape tracking with the Full method and whether a multiplier selected for one robot is also suitable for other robots.
Using the same FR3 sequence, initial configurations, and control
settings as in the first experiment, we set
$\bm M_d^{(c)}(t)=c\bm M_{\mathrm{FR3}}(t)$.
We sweep constant multipliers $c>0$ and select, for each robot, the
candidate minimizing $d_s$ RMS during motion ($3$--$8$~s). 
Because each multiplier is chosen from the follower's own realized trajectory, it is the most favorable constant scale available to the Full method in this comparison; a transfer pipeline that estimates the factor from the demonstration alone does not have this information.
The selected values are $1.189$, $11.31$, and $5.657$ for Gen3,
KR500, and UR20, respectively.
Each multiplier is applied to all three followers, with the
results from the Shape method from the first experiment used as the reference.

We evaluate the semi-axis ratio
$r=\sqrt{\lambda_{\max}/\lambda_{\min}}$, using the largest
and smallest eigenvalues of each ME to obtain the current
ratio $r$ and target ratio $r_d$.
The major-axis error is $\theta=\arccos(|\bm u_c^\top\bm u_d|)$
in degrees, where $\bm u_c$ and $\bm u_d$ are unit eigenvectors
associated with the largest eigenvalues of the current and target MEs.
Positive scaling preserves the target ratio and principal directions.

With each robot's own selected multiplier, the Full method yields
motion-phase mean shape distances $\overline d_s=0.2210$--$0.2221$,
close to approximately $0.2210$ with the Shape method
(Fig.~\ref{fig:target-scale-sensitivity}(a)).
Applying the Gen3-selected multiplier instead gives
$\overline d_s=1.6470$ for KR500 and $0.7410$ for UR20.

Table~\ref{tab:scale-errors} reports the endpoint errors
$|r(11)-r_d(11)|$, $\theta(11)$, and $d_s(11)$,
with $r_d(11)=4.0344$; ratio and shape errors are dimensionless.
The Shape method achieves $d_s(11)\approx9.30\times10^{-5}$
for all three followers (Sec.~\ref{sec:robot_transfer_experiment}),
whereas the Full method retains $d_s(11)=0.0684$ for KR500
and $0.1113$ for UR20 even with their own selected multipliers.
For these two robots, axis-ratio errors remain despite small
major-axis errors, showing that major-axis alignment can
coexist with shape mismatch.

Fig.~\ref{fig:target-scale-sensitivity}(b) decomposes the squared
total AIRM distance relative to each scaled target $\bm M_d^{(c)}$
using \eqref{eq:scale_shape_distance}.
For the six combinations using a multiplier selected for a different
robot, the final-hold ratio
$\overline{d_\rho^2}/\overline{d_{\mathrm{AI}}^2}$ is
$66.4$--$97.9\%$, indicating that scale differences dominate
the squared total distance in these cases.

Thus, the effectiveness of constant target scaling depends
on both the robot and evaluation phase, whereas the Shape method
tracks orientation and semi-axis length ratios without selecting
a target-scale multiplier.

\begin{table}[t]
  % Local caption layout; numbering and references still use \caption.
  \makeatletter
  \long\def\@makecaption#1#2{%
    {\centering\footnotesize #1.\enspace{\scshape #2}\par}%
    \vskip 5pt}
  \makeatother
  \caption{Human-transfer errors}
  \label{tab:human-errors}
  \centering
  \footnotesize
  \setlength{\tabcolsep}{2pt}
  \renewcommand{\arraystretch}{1.15}
  % Natural column widths reserve space for each header and right-aligned value.
  \begin{tabular*}{\columnwidth}{@{\extracolsep{\fill}}cl|rrr|rrr@{}}
    % Arithmetic means over disjoint phases: 2,500 / 620 / 1,501 states.
    % Source: analysis/human_reach/phase_review_20260913/METRICS.csv
    \hline
    \multicolumn{8}{c}{Phase means} \\
    \hline
    &
    & \multicolumn{3}{c|}{Full} & \multicolumn{3}{c}{Shape} \\
    \hline
    \rule{0pt}{2.5ex}\raisebox{-0.5\baselineskip}[0pt][0pt]{Robot}
    & \raisebox{-0.5\baselineskip}[0pt][0pt]{Phase [s]}
    & \multicolumn{1}{c}{$\overline{e}_p$}
    & \multicolumn{1}{c}{$\overline{\theta}_{\mathrm{dir}}$}
    & \multicolumn{1}{c|}{$\overline{d}_s$}
    & \multicolumn{1}{c}{$\overline{e}_p$}
    & \multicolumn{1}{c}{$\overline{\theta}_{\mathrm{dir}}$}
    & \multicolumn{1}{c}{$\overline{d}_s$} \\[-1pt]
    & & \multicolumn{1}{c}{$[\%L_r]$}
    & \multicolumn{1}{c}{$[\mathrm{deg}]$} & \multicolumn{1}{c|}{}
    & \multicolumn{1}{c}{$[\%L_r]$}
    & \multicolumn{1}{c}{$[\mathrm{deg}]$} & \\
    \hline
    FR3 & $[0,5)$ & 13.251 & 17.154 & 0.640 & 6.607 & 12.332 & 0.595 \\
     & $[5,6.24)$ & 18.341 & 20.085 & 0.741 & 10.695 & 8.451 & 0.480 \\
     & $[6.24,9.24]$ & 8.123 & 25.649 & 0.423 & 5.507 & 3.420 & 0.312 \\
    \hline
    Gen3 & $[0,5)$ & 15.965 & 25.989 & 0.595 & 4.754 & 18.834 & 0.469 \\
     & $[5,6.24)$ & 26.731 & 30.964 & 0.944 & 9.422 & 16.689 & 0.503 \\
     & $[6.24,9.24]$ & 13.362 & 40.725 & 0.598 & 4.064 & 5.667 & 0.237 \\
    \hline
    KR500 & $[0,5)$ & 35.531 & 62.890 & 0.936 & 6.371 & 9.785 & 0.582 \\
     & $[5,6.24)$ & 59.707 & 78.085 & 0.579 & 9.396 & 9.002 & 0.605 \\
     & $[6.24,9.24]$ & 75.103 & 78.423 & 1.205 & 6.156 & 2.644 & 0.337 \\
    \hline
    UR20 & $[0,5)$ & 32.314 & 20.482 & 1.160 & 4.533 & 20.952 & 0.399 \\
     & $[5,6.24)$ & 49.289 & 11.186 & 1.361 & 7.314 & 4.965 & 0.505 \\
     & $[6.24,9.24]$ & 66.884 & 10.875 & 0.941 & 3.980 & 2.103 & 0.267 \\
    \hline
  \end{tabular*}
  \par\nointerlineskip
  % Equal metric-column widths center method headers between group boundaries.
  % Inset right-aligned endpoint values by 7pt within each metric column.
  \begin{tabularx}{\columnwidth}{@{}c|*{3}{>{\raggedleft\arraybackslash\rightskip=7pt\relax}X}|*{3}{>{\raggedleft\arraybackslash\rightskip=7pt\relax}X}@{}}
    \multicolumn{7}{c}{Endpoint ($t=9.24$~s)} \\
    \hline
    & \multicolumn{3}{c|}{Full} & \multicolumn{3}{c}{Shape} \\
    \hline
    \rule{0pt}{2.5ex}\raisebox{-0.5\baselineskip}[0pt][0pt]{Robot}
    & \multicolumn{1}{c}{$e_p$}
    & \multicolumn{1}{c}{$\theta_{\mathrm{dir}}$}
    & \multicolumn{1}{c|}{$d_s$}
    & \multicolumn{1}{c}{$e_p$}
    & \multicolumn{1}{c}{$\theta_{\mathrm{dir}}$}
    & \multicolumn{1}{c}{$d_s$} \\[-1pt]
    & \multicolumn{1}{c}{$[\%L_r]$}
    & \multicolumn{1}{c}{$[\mathrm{deg}]$} & \multicolumn{1}{c|}{}
    & \multicolumn{1}{c}{$[\%L_r]$}
    & \multicolumn{1}{c}{$[\mathrm{deg}]$} & \\
    \hline
    FR3 & 4.782 & 31.424 & 0.368 & 4.289 & 4.356 & 0.322 \\
    Gen3 & 6.134 & 37.450 & 0.477 & 2.428 & 3.713 & 0.193 \\
    KR500 & 75.104 & 78.426 & 1.205 & 5.603 & 1.564 & 0.304 \\
    UR20 & 67.018 & 10.826 & 0.863 & 3.360 & 2.334 & 0.222 \\
    \hline
  \end{tabularx}
  \vspace{-0.5cm}
\end{table}

\subsection{Human-to-Robot Transfer with Concurrent Tasks}
\label{sec:human_transfer_experiment}

The third experiment evaluates whether human ME geometry can
be transferred to four heterogeneous manipulators while tracking
EE position and direction.
We selected a forward reach from a preparatory posture because
the recorded motion combines wrist displacement with increasing
elongation of the dual force ME along world $+X$, providing
a reference for transferring both motion and directional geometry.
We used a $1.24$~s segment of right-arm motion-capture data
recorded at $100$~Hz.
Floating-base positions and joint angles were interpolated
linearly, and floating-base rotations by quaternion spherical linear interpolation;
wrist positions and MEs were recomputed at $500$~Hz.
The sequence comprises an initial hold over $[0,5)$,
reaching over $[5,6.24)$, and a final hold over $[6.24,9.24]$~s.

The human ME was computed as $\bm M_h=\bm J_h\bm J_h^\top$,
using the world-frame translational wrist Jacobian restricted
to six shoulder and forearm Cardan coordinates.
The position reference was
$\bm p_d=\bm o_r+(L_r/L_h)(\bm p_{h,w}-\bm p_{h,s})$,
where $\bm p_{h,w}$ and $\bm p_{h,s}$ are the human wrist and
shoulder positions in the world frame, $\bm o_r$ is a fixed
robot reference point, and $L_h$ is the sum of the human upper-arm
and forearm lengths.
Using all arm joints, we applied \eqref{eq:joint_velocity_qp}
with position and direction tasks.
The position residual was normalized by $L_r$.
The direction task aligned the EE $+Z$ unit direction $\bm a_c$,
expressed in the world frame, with the world $+X$ unit vector
$\bm e_X$, leaving roll free.
The position, direction, and ME weights were $(20,0.5,1)$,
with feedback gains $(4,4,3)$ and $\eta=0.002$.
Neither method received a robot-specific target scale; Section~\ref{sec:target_scale_experiment} characterizes the calibrated case for the Full method, where a suitable multiplier has to be fitted to each robot and the resulting choice depends on the phase used to fit it.

Position and direction errors were evaluated as
$e_p=100\|\bm p_c-\bm p_d\|_2/L_r$ and
$\theta_{\mathrm{dir}}=\arccos(\bm a_c^\top\bm e_X)$,
where $\bm p_c$ is the current EE position.
The errors are reported in percent of $L_r$ and degrees,
respectively.
Fig.~\ref{fig:human-transfer-full-shape-xz-yz}(a) shows
EE paths and error histories, while Table~\ref{tab:human-errors}
reports phase means $\overline e_p$,
$\overline\theta_{\mathrm{dir}}$, and $\overline d_s$,
and endpoint values.
At the endpoint, the dual force shapes of all four robots
using the Shape method are elongated along $X$, as in the
human target
(Figs.~\ref{fig:human-to-robot-shape-motion}
and~\ref{fig:human-transfer-full-shape-xz-yz}(b)).

During reaching, the Shape method yields lower
$\overline e_p$ and $\overline\theta_{\mathrm{dir}}$
than the Full method for all four robots,
and lower $\overline d_s$ except for KR500.
During the final hold and at the endpoint, all three metrics
are lower with the Shape method for all four robots.
The endpoint position error for KR500 is $75.104\%$ of $L_r$
with the Full method and $5.603\%$ with the Shape method;
for UR20, the corresponding errors are $67.018\%$ and $3.360\%$.

In this experiment, with no target-scale selection for either method, the Shape method reproduces the human's directional geometry with smaller position and EE direction errors. The errors of the Full method reflect both the scale mismatch quantified in Section~\ref{sec:target_scale_experiment} and the competing position task; they characterize uncalibrated transfer rather than full-matrix tracking in general.

\section{Conclusion}

We presented a scale-invariant manipulability shape-tracking
method on the space of unit-determinant SPD matrices,
transferring ellipsoid orientation and semi-axis length ratios
across systems with different sizes and kinematic structures.
Using the scale--shape decomposition, we separated the scale
component from the AIRM tracking cost and constructed an
orthonormal residual representation that preserves the shape
cost.
The resulting objective requires no absolute-scale matching
and integrates with position and EE direction tasks
in a constrained joint-velocity QP.

In kinematic simulations, the Shape method reproduced a common
target shape on three robot followers with small endpoint errors
without robot-specific target-scale adjustment.
Robot-specific scaling brought the Full method's shape-tracking
performance close to that of the Shape method during motion,
but its effectiveness depended on the robot and evaluation phase.
For the recorded human reach, the Shape method reproduced the
forward elongation of the target dual force shape while tracking
EE position and direction.
Without robot-specific ME scale calibration, all four robots
had smaller position, direction, and shape errors with the
Shape method than with the Full method during the final hold
and at the endpoint.
These results support the proposed method for transferring
ME geometry alongside concurrent tasks without selecting
an absolute target scale.

The current evaluation is limited to kinematic simulations, focusing on the transfer of manipulability through robot motion. Future work will evaluate manipulability transfer on physical
robots during task execution, including dynamics and contact
interactions under kinematic constraints and concurrent tasks. In addition, we will extend the current evaluation beyond a single human reaching motion to a broader range of complex human motions, with the goal of examining manipulability transfer in the broader context of human-to-robot skill transfer.


%%%%%%%%%%%%%%%%%%%%%%%%%%%%%%%%%%%%%%%%%%%%%%%%%%%%%%%%%%%%%%%%%%%%%%%%%%%%%%%%

%%%%%%%%%%%%%%%%%%%%%%%%%%%%%%%%%%%%%%%%%%%%%%%%%%%%%%%%%%%%%%%%%%%%%%%%%%%%%%%%

%%%%%%%%%%%%%%%%%%%%%%%%%%%%%%%%%%%%%%%%%%%%%%%%%%%%%%%%%%%%%%%%%%%%%%%%%%%%%%%%
\useRomanappendicesfalse
\appendices
\section{}
\label{app:shape_coordinates}

Let $\bm{E}_{ij}$ be the $D\times D$ matrix with its sole
nonzero entry equal to one at $(i,j)$.
Dividing the $m-1$ real symmetric generalized Gell--Mann
matrices~\cite{bertlmann2008bloch} by $\sqrt{2}$ gives a
Frobenius-orthonormal basis of $\operatorname{Sym}_0(D)$,
indexed by $\alpha$:
\begin{equation}
\bm{B}_{\ell}^{\mathrm{diag}}
=
\frac{
\sum_{i=1}^{\ell}\bm{E}_{ii}
-\ell\bm{E}_{\ell+1,\ell+1}
}{\sqrt{\ell(\ell+1)}},
\quad 1\leq\ell\leq D-1,
\label{eq:app_diagonal_basis}
\end{equation}
\begin{equation}
\bm{B}_{ij}^{\mathrm{off}}
=
\frac{\bm{E}_{ij}+\bm{E}_{ji}}{\sqrt{2}},
\quad 1\leq i<j\leq D.
\label{eq:app_offdiagonal_basis}
\end{equation}
For $\bm{X}\in\operatorname{Sym}_0(2)$, placing the diagonal
basis matrix first gives
\begin{equation}
\operatorname{vec}_0^{(2)}(\bm{X})
=
\begin{bmatrix}
(X_{11}-X_{22})/\sqrt{2}\\
\sqrt{2}X_{12}
\end{bmatrix}.
\label{eq:app_shape_coordinates_2d}
\end{equation}
For $\bm{X}\in\operatorname{Sym}_0(3)$, use diagonal order
$\ell=1,2$ and off-diagonal order $(1,2)$, $(2,3)$, $(1,3)$:
\begin{equation}
\operatorname{vec}_0^{(3)}(\bm{X})
=
\begin{bmatrix}
(X_{11}-X_{22})/\sqrt{2}\\
(X_{11}+X_{22}-2X_{33})/\sqrt{6}\\
\sqrt{2}X_{12}\\
\sqrt{2}X_{23}\\
\sqrt{2}X_{13}
\end{bmatrix}.
\label{eq:app_shape_coordinates_3d}
\end{equation}

Let $\bm{P}=\widehat{\bm{M}}_c$ be the current shape.
Differentiating $\log\det\widehat{\bm{M}}=0$
with respect to $q_j$ gives
$\operatorname{tr}
(\bm{P}^{-1}\widehat{\bm{\mathcal{J}}}_{:,:,j})=0$.
The relative matrix
$\bm{R}=\bm{P}^{-1/2}\widehat{\bm{M}}_d\bm{P}^{-1/2}$
is SPD and satisfies $\det\bm{R}=1$.
Hence the logarithm map satisfies
$\operatorname{tr}
(\bm{P}^{-1}\operatorname{Log}_{\bm{P}}
(\widehat{\bm{M}}_d))
=\operatorname{tr}(\log\bm{R})=0$.
Thus the residual
$\bm{U}=\dot{\widehat{\bm{M}}}_c
-k_M\operatorname{Log}_{\bm{P}}(\widehat{\bm{M}}_d)$
belongs to $T_{\bm{P}}\mathcal{SP}(D)$, and
$\bm{X}=\bm{P}^{-1/2}\bm{U}\bm{P}^{-1/2}$
is symmetric and traceless.

At fixed $\bm{P}$, linearity gives
$\mathcal C_{\bm{P}}(\bm{U})
=\bm{J}_M\dot{\bm{q}}-k_M\bm{e}_s$.
Orthonormality and \eqref{eq:airm_tangent_norm} yield
\begin{equation}
\begin{aligned}
\|\mathcal C_{\bm{P}}(\bm{U})\|_2^2
&=
\sum_{\alpha=1}^{m-1}
\left[\operatorname{tr}(\bm{B}_{\alpha}\bm{X})\right]^2
=
\|\bm{X}\|_F^2
=
\|\bm{U}\|_{\bm{P}}^2.
\end{aligned}
\label{eq:app_airm_norm_preservation}
\end{equation}
Multiplying by $1/2$ proves the equivalence of
\eqref{eq:shape_tracking_cost} and
\eqref{eq:shape_tracking_cost_airm}.

%%%%%%%%%%%%%%%%%%%%%%%%%%%%%%%%%%%%%%%%%%%%%%%%%%%%%%%%%%%%%%%%%%%%%%%%%%%%%%%%

\bibliographystyle{IEEEtran}
\bibliography{ref} % Entries are in the refs.bib file

@article{yoshikawa1985manipulability,
  title={Manipulability of robotic mechanisms},
  author={Yoshikawa, Tsuneo},
  journal={The international journal of Robotics Research},
  volume={4},
  number={2},
  pages={3--9},
  year={1985},
  publisher={Sage Publications Sage CA: Thousand Oaks, CA}
}

@inproceedings{jaquier2018geometry,
  title={Geometry-aware Tracking of Manipulability Ellipsoids.},
  author={Jaquier, No{\'e}mie and Rozo, Leonel Dario and Caldwell, Darwin G and Calinon, Sylvain},
  booktitle={Robotics: Science and Systems},
  number={CONF},
  year={2018}
}

@inproceedings{jaquier2020analysis,
  title={Analysis and transfer of human movement manipulability in industry-like activities},
  author={Jaquier, No{\'e}mie and Rozo, Leonel and Calinon, Sylvain},
  booktitle={2020 IEEE/RSJ International Conference on Intelligent Robots and Systems (IROS)},
  pages={11131--11138},
  year={2020},
  organization={IEEE}
}

@inproceedings{rozo2017learning,
  title={Learning manipulability ellipsoids for task compatibility in robot manipulation},
  author={Rozo, Leonel and Jaquier, No{\'e}mie and Calinon, Sylvain and Caldwell, Darwin G},
  booktitle={2017 IEEE/RSJ International Conference on Intelligent Robots and Systems (IROS)},
  pages={3183--3189},
  year={2017},
  organization={IEEE}
}

@article{yan2021decentralized,
  title={Decentralized ability-aware adaptive control for multi-robot collaborative manipulation},
  author={Yan, Lei and Stouraitis, Theodoros and Vijayakumar, Sethu},
  journal={IEEE Robotics and Automation Letters},
  volume={6},
  number={2},
  pages={2311--2318},
  year={2021},
  publisher={IEEE}
}

@article{abu2021probabilistic,
  title={A probabilistic framework for learning geometry-based robot manipulation skills},
  author={Abu-Dakka, Fares J and Huang, Yanlong and Silv{\'e}rio, Jo{\~a}o and Kyrki, Ville},
  journal={Robotics and Autonomous Systems},
  volume={141},
  note={{Art. no.} 103761},
  year={2021},
  publisher={Elsevier}
}

@article{choi2023simple,
  title={Simple desired manipulability ellipsoid with velocity and force for control of redundant manipulator},
  author={Choi, Yun Seok and Rhee, Issac and Hoang, Phi Tien and Choi, Hyouk Ryeol},
  journal={Journal of Mechanical Science and Technology},
  volume={37},
  number={4},
  pages={2033--2041},
  year={2023},
  publisher={Springer}
}

@article{jaquier2021geometry,
  title={Geometry-aware manipulability learning, tracking, and transfer},
  author={Jaquier, No{\'e}mie and Rozo, Leonel and Caldwell, Darwin G and Calinon, Sylvain},
  journal={The International Journal of Robotics Research},
  volume={40},
  number={2--3},
  pages={624--650},
  year={2021},
  publisher={Sage Publications Sage UK: London, England}
}

@article{abu2024unified,
  title={A unified formulation of geometry-aware discrete dynamic movement primitives},
  author={Abu-Dakka, Fares J and Saveriano, Matteo and Kyrki, Ville},
  journal={Neurocomputing},
  volume={598},
  pages={128056},
  year={2024},
  publisher={Elsevier}
}

@inproceedings{li2025manidp,
  title={ManiDP: Manipulability-aware diffusion policy for posture-dependent bimanual manipulation},
  author={Li, Zhuo and Liu, Junjia and Li, Dianxi and Teng, Tao and Li, Miao and Calinon, Sylvain and Caldwell, Darwin and Chen, Fei},
  booktitle={2025 IEEE/RSJ International Conference on Intelligent Robots and Systems (IROS)},
  pages={9956--9962},
  year={2025},
  organization={IEEE}
}

@inproceedings{reithmeir2022human,
  title={Human-to-robot manipulability domain adaptation with parallel transport and manifold-aware ICP},
  author={Reithmeir, Anna and Figueredo, Luis and Haddadin, Sami},
  booktitle={2022 IEEE/RSJ International Conference on Intelligent Robots and Systems (IROS)},
  pages={5218--5225},
  year={2022},
  organization={IEEE}
}

@inproceedings{gong2025manipulability,
  title={Manipulability Transfer and Tracking Control: Bridging Domain Adaptation with Predictive Feasibility},
  author={Gong, Yuhe and Xing, Hao and Yu, Guo and Figueredo, Luis},
  booktitle={2025 IEEE International Conference on Robotics and Automation (ICRA)},
  pages={4572--4578},
  year={2025},
  organization={IEEE}
}

@article{moakher2005differential,
  title={A differential geometric approach to the geometric mean of symmetric positive-definite matrices},
  author={Moakher, Maher},
  journal={SIAM journal on matrix analysis and applications},
  volume={26},
  number={3},
  pages={735--747},
  year={2005},
  publisher={SIAM}
}

@article{bertlmann2008bloch,
  title={Bloch vectors for qudits},
  author={Bertlmann, Reinhold A and Krammer, Philipp},
  journal={Journal of Physics A: Mathematical and Theoretical},
  volume={41},
  number={23},
  pages={235303},
  year={2008}
}

@article{chiu1988task,
  title={Task compatibility of manipulator postures},
  author={Chiu, Stephen L},
  journal={The international journal of robotics research},
  volume={7},
  number={5},
  pages={13--21},
  year={1988},
  publisher={Sage Publications Sage CA: Thousand Oaks, CA}
}

@article{ajoudani2017choosing,
  title={Choosing poses for force and stiffness control},
  author={Ajoudani, Arash and Tsagarakis, Nikos G and Bicchi, Antonio},
  journal={IEEE Transactions on Robotics},
  volume={33},
  number={6},
  pages={1483--1490},
  year={2017},
  publisher={IEEE}
}

@article{pennec2006riemannian,
  title={A Riemannian framework for tensor computing},
  author={Pennec, Xavier and Fillard, Pierre and Ayache, Nicholas},
  journal={International Journal of computer vision},
  volume={66},
  number={1},
  pages={41--66},
  year={2006},
  publisher={Springer}
}

@article{dolcetti2019differential,
  author  = {Alberto Dolcetti and Donato Pertici},
  title   = {Differential Properties of Spaces of Symmetric Real Matrices},
  journal = {Rendiconti del Seminario Matematico, Universit{\`a} e Politecnico di Torino},
  year    = {2019},
  volume  = {77},
  number  = {1},
  pages   = {25--43}
}

\end{document}